\documentclass[letterpaper, 10 pt, journal, twoside]{ieeetran}
\usepackage{cite}
\usepackage{amsmath,amssymb,amsfonts}
\usepackage{algorithmic}
\usepackage{graphicx}
\usepackage{textcomp}
\usepackage{xcolor}
\usepackage{amsthm}
\usepackage[ruled,vlined]{algorithm2e}
\usepackage{comment}
\usepackage{caption}
\usepackage{subcaption}
\usepackage{array}
\usepackage{multirow}
\usepackage{lipsum}
\usepackage{hhline}

\def\BibTeX{{\rm B\kern-.05em{\sc i\kern-.025em b}\kern-.08em
    T\kern-.1667em\lower.7ex\hbox{E}\kern-.125emX}}

\begin{document}

\title{Trust-Preserved Human-Robot Shared Autonomy enabled by Bayesian Relational Event Modeling}

\author{Yingke Li$^{1}$, and Fumin Zhang$^{1,2}$%
\thanks{Manuscript received: February 22, 2024; Revised May 21, 2024; Accepted July 12, 2024.}
\thanks{This paper was recommended for publication by Editor Angelika Peer upon evaluation of the Associate Editor and Reviewers' comments.}
\thanks{This work was supported by ONR grants N00014-19-1-2556 and N00014-19-1-2266; AFOSR grant FA9550-19-1-0283 and FA9550-22-1-0244; NSF grants GCR-1934836, CNS-2016582, DMS-2053489 and ITE-2137798; and NOAA grant NA16NOS0120028.} 
\thanks{$^{1}$Yingke Li and Fumin Zhang are with School of Electrical and Computer Engineering, Georgia Institute of Technology, Atlanta, GA, USA
        {\tt\footnotesize yli3225, fumin@gatech.edu}}
\thanks{$^{2} $Fumin Zhang is with Department of Electronic and Computer Engineering, Hong Kong University of Science and Technology, Kowloon, Hong Kong, China
        {\tt\footnotesize eefumin@ust.hk}}
\thanks{Digital Object Identifier (DOI): see top of this page.}
\thanks{The user study in this paper was found to pose minimal risk and has been determined to be Exempt by the Georgia Tech IRB (Protocol Number: H23412). This determination is effective as of 01/10/2024.}
}

\maketitle

\begin{abstract}
\textit{Shared autonomy} functions as a flexible framework that empowers robots to operate across a spectrum of autonomy levels, allowing for efficient task execution with minimal human oversight.
However, humans might be intimidated by the autonomous decision-making capabilities of robots due to perceived risks and a lack of trust.
This paper proposed a trust-preserved shared autonomy strategy that allows robots to seamlessly adjust their autonomy level, striving to optimize team performance and enhance their acceptance among human collaborators.
By enhancing the \textit{relational event modeling} framework with Bayesian learning techniques, this paper enables dynamic inference of human trust based solely on time-stamped relational events communicated within human-robot teams.
Adopting a longitudinal perspective on trust development and calibration in human-robot teams, the proposed trust-preserved shared autonomy strategy warrants robots to actively establish, maintain, and repair human trust, rather than merely passively adapting to it.
We validate the effectiveness of the proposed approach through a user study on a human-robot collaborative search and rescue scenario.
The objective and subjective evaluations demonstrate its merits on both task execution and user acceptability over the baseline approach that does not consider the preservation of trust.
\end{abstract}

\begin{IEEEkeywords}
Acceptability and Trust, Human-Robot Teaming, Probability and Statistical Methods
\end{IEEEkeywords}

\section{Introduction}

\IEEEPARstart{A}{s} robots continue to find applications in a wide range of real-world scenarios, the next wave of technological progress in the field of robotics appears to be centered around their ability to interact effectively with complex environments and autonomously execute tasks with minimal human supervision \cite{Selvaggio2021AutonomySurvey}. 
Although maintaining human global control over autonomous systems contributes to increasing the initiative and awareness of dynamic, complex, and interactive environments, the ability of robots to make unaided decisions also grows in significance \cite{Liu2022CoordinatingProficiencies}.

On the one hand, the efficiency of the human-robot teams can be compromised due to frequent human intervention \cite{Beer2014TowardInteraction, Franchi2012BilateralTopology}. 
From the human perspective, overseeing and instructing robots, particularly in multi-robot teams where a large group of robots require simultaneous management, can demand a significant commitment of effort.
For robots, continuous interruptions from human can cause disturbances that make their operation destabilized, dependent, and slow \cite{Schermerhorn2009DynamicTask}, especially in the presence of communication delays.

\begin{figure}
    \centering
    \includegraphics[width=0.46\textwidth]{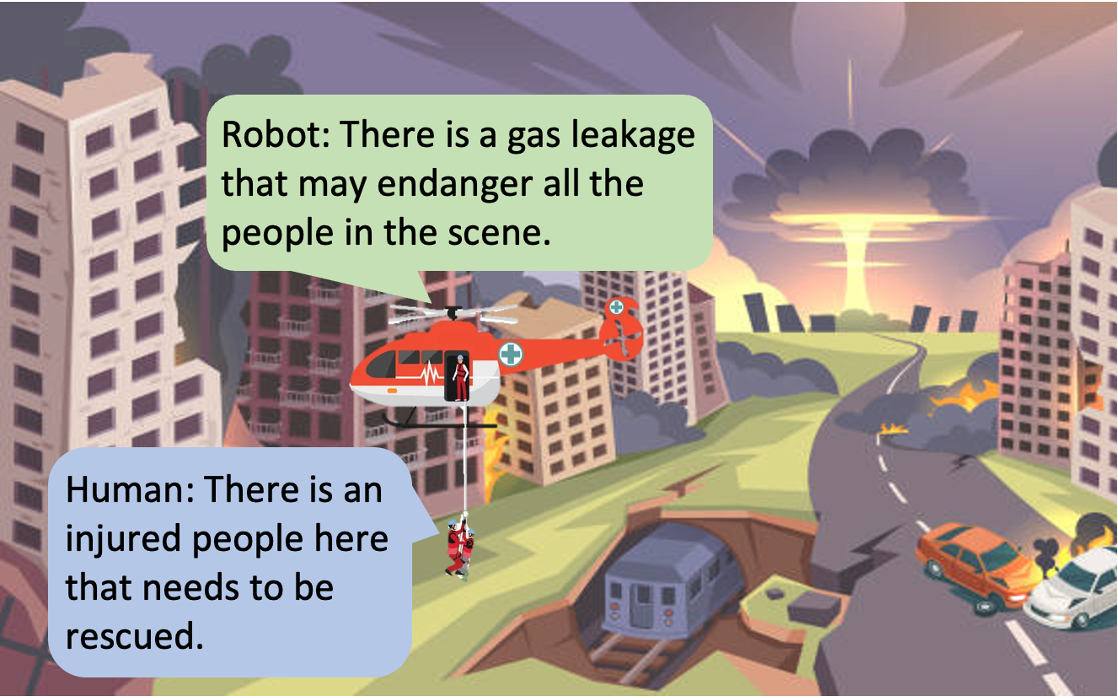}
    \caption{A search and rescue scenario with human-robot team. Robots may acquire information unavailable to human team members, and make strictly rational decisions.}
    \label{fig:SAR}
\end{figure}

On the other hand, the fundamental expectation behind human-robot teaming is that the abilities of robots and humans complement each other, resulting in enhanced performance compared to their individual efforts \cite{Angleraud2021CoordinatingCommands}. 
For example, robots can often offer an information advantage by accessing data that are unavailable to human team members \cite{Schermerhorn2009DynamicTask}.
Consider a human-robot collaborative search and rescue (SAR) scenario as illustrated in Fig. \ref{fig:SAR}. 
The human responder might initially focus on rescuing injured individuals. 
At the same time, the robot identifies a gas leakage that requires immediate attention to prevent a potential explosion endangering everyone within the scene. 
In such a situation, the robot is challenged to weigh the importance of two conflicting goals: adhering to the human's instructions to continue rescuing the victim versus prioritizing putting down the gas leakage.
Hence in the best interests of the team as a whole, the robots are expected to make judicious and rational assessments to determine when to adhere to or disregard human directives when conflicts happen.

The degree to which the robots autonomously choose to follow human instructions in the light of other objectives can be characterized by their autonomy level \cite{Beer2014TowardInteraction}. 
\textit{Shared autonomy} \cite{Mostafa2019AdjustableReview}, also referred to as \textit{adaptive autonomy}, provides an opportunity to establish a flexible autonomy framework that grants robots the capacity to function across varying autonomy levels. 
However, the autonomous decision-making capabilities of robotic systems can sometimes be unsettling for humans due to various factors, including perceived risks and a lack of trust \cite{Desai2009CreatingSystems}. 
Therefore, achieving fluent human-robot teaming mandates addressing concerns regarding the acceptability of non-human entities making decisions autonomously on behalf of human team members. 
As such, this paper pays special attention to maintaining human trust during the adaptation of the robot's autonomy level.

Motivated by the shared properties and constraints of real-world human-robot teaming where humans and robots may be spatially and temporally separated, we investigate a simple but representative scenario of human-robot collaborative SAR \cite{Schoonderwoerd2022DesignTask}.
In this scenario, humans and robots have complementary capabilities, and tasks are dependent on sequences, requiring collaboration among different team members.
In addition, since the robots are separated from their remote human operators in both space and time, their communication are restricted to merely time-stamped commands communicated between them.

To captivate trust-preserved shared autonomy within such challenging circumstances, the biggest obstacle comes from the inherent difficulty of inferring human trust from solely a ﬂow of commands communicated within the human-robot team, where none of these events by themselves characterize the team relationships at any point in time.
To tackle this challenge, we propose a novel approach to dynamically infer the underlying drivers of interaction dynamics (e.g. trust) from merely time-stamped relational events (e.g. commands) by enhancing the relational event modeling (REM) framework \cite{Schecter2021TheModel, Mulder2019ModelingAnalysis} with Bayesian learning techniques.
The human trust inferred online is thus an evolving process, which allows us to take a longitudinal perspective on trust development and calibration in human-robot teams \cite{DeVisser2020TowardsTeams}.
Leveraging the punctuated equilibrium model (PEM) \cite{Sabherwal2001TheModel}, we developed a trust-preserved shared autonomy strategy that allows the robots to seamlessly adjust their autonomy levels while actively establishing, maintaining, and repairing human trust.

The contributions of this paper are summarized as:

\noindent \textbf{Trust Inference from Remote Interactions.} We propose an innovative approach to dynamically infer human trust based on remote human-robot interactions.

\noindent \textbf{Trust-preserved Shared Autonomy.} We devise a trust-preserved shared autonomy strategy that maintains human trust during the adaptation of robot's autonomy level.

\noindent \textbf{User Study on Collaborative SAR.} We validate the effectiveness of the proposed method by performing a user study on a human-robot collaborative SAR scenario.

\section{Related Work}

\noindent \textbf{Trust Inference.}
The priority of trustworthy human-robot teaming is to capture a relatively accurate understanding of human trust.
In the field of human-robot interaction, numerous researchers have approached this challenge from diverse viewpoints, for example, physiological measures (e.g. skin conductance, eye-tracking), neuroimaging, behavioral observations, and self-reported trust ratings \cite{Khalid2020DeterminantsPredicting, Guo2020ModelingApproach, Zheng2018Human-RobotSystems}.
Those approaches function effectively in the scenarios of proximate human-robot interaction where humans and robots share the same workspace, allowing for ample human observations. 
However, the majority of those methods lose their throne in more general real-world scenarios of human-robot teaming where humans and robots are separated spatially and temporally.
This paper opens new possibilities of rigorously inferring human trust from remote human-robot interactions, which possesses the potential to be applied to more general and complicated real-world human-robot teaming scenarios.

\noindent \textbf{Shared Autonomy (SA).}
Shared autonomy approaches allow the robot to seamlessly adapt its autonomy level based on its own understanding of the human actions/intentions and of the surrounding environment.
There have been various autonomy adaptation strategies proposed utilizing information extracted from both the human (e.g. intentions, muscle activity, skills) and the environment (e.g. task, constraints) \cite{Li2015ContinuousControl, Milliken2017ModelingAutonomy, Enayati2018Skill-basedTracking, Jain2020ProbabilisticRobotics}.
Although effective to a certain extent, few of those approaches take into account an essential fact that real-world human-robot teaming is a long-term and evolutionary process.
Instead, this paper takes a longitudinal perspective on trust development and calibration in human-robot teams, and devises a trust-preserved shared autonomy strategy that actively maintains human trust during the adaptation of robot’s autonomy level.

\section{Background on Relational Event Modelling}

\subsection{Relational Event Modelling}\label{subsec:REM}

Interactions among individuals can be represented as a sequence of \textit{relational event}, which is defined as ``a discrete event generated by a social actor and directed toward one or more targets'' \cite{Butts2017ADynamics}. 
Formally, a relational event is represented by a tuple $e=(i_{e},j_{e},k_{e},w_{e},t_{e})$ containing the sender $i_{e}$, receiver $j_{e}$, type $k_{e}$, weight $w_{e}$, and time $t_{e}$ of an event.
Let $A=\{1,\cdots,n\}$ be the set of all $n$ actors, and $D \subseteq A \times A$ the set of all dyads. 
Events can fall into a set of $K$ discrete classes.
For each class $k$, define the weight function $w_{kt}$ as the accumulated weight of all past interaction events \cite{Brandes2009NetworksData}.

Such events constitute a dynamic network of directional
ties, where the probability of an event occurring often depends on various kinds of attributes in the network. 
For example, individual covariates such as gender, age, or institutional role; dyadic covariates such as friendship status or trust; environmental factors including information about the physical environment, the stage in a creative process, or the status of an emergency.
For simplicity of notation, we combine all those attribute information into a set $G_{t}$.

The full sequence of $m$ events formulates the history event set $E=\{e_{1},\cdots,e_{m}\}$. 
The occurrence of future relational events can be influenced by both the history of past events $E$ and the attributes within the network $G_t$.
REM is a statistical technique that blends social sequence and network analysis, which provides a principled approach to investigate the underlying factors that drive the occurrence, rhythm, and speed of individuals interacting with each other over time.
REM models every potential pairing event of individuals in terms of rates, where common events have high rates, and rare events have low rates.

Following the event history approach \cite{Blossfeld2001TechniquesAnalysis}, the probability of the sequence and timing of events can be defined by the hazard rate and survival function for each possible action.
The hazard rate can be interpreted as the instantaneous likelihood of an event occurring, given that it has not yet occurred. 
The survival function can be interpreted as the likelihood of an event occurring.
Survival models relate the time that passes, before some event occurs, to one or more covariates that may be associated with that quantity of time.
Proportional hazards models are a class of survival models assuming that, the unique effect of a unit increase in a covariate is multiplicative with respect to the hazard rate, which allows us to estimate the effect parameters without any consideration of the full hazard function.
Therefore, REM utilizes the \textit{Cox proportional hazards model} \cite{Cox1972RegressionLifeTables}, where the hazard rate for a sender $i$ directing an action of type $k$ to receiver $j$ at time $t$ is modeled as a log-linear function of a set of sufficient statistics $S_{p}(i,j,k,G_{t})$ and their corresponding intensity parameters $\theta_{p}$:
\begin{equation*}
    \lambda_{ijk}(t|G_{t}, \boldsymbol{\theta}) = \lambda_{0}(t) \exp{\left(\sum_{p=1}^{P} \theta_{p} S_{p}(i,j,k,G_{t})\right)},
\end{equation*}
where $\boldsymbol{\theta}=[\theta_{1}, \cdots, \theta_{P}]^{\intercal}$,  and $p \in \{1,\cdots,P\}$ is the index of the sufficient statistics,
and the parameter $\lambda_{0}(t)$ describes how the risk of event per time unit changes over time at baseline levels of covariates. 

The likelihood function for the full event sequence and timing can be explicitly computed by combining the hazard rate for each realized event with the survival function for all possible events. 
When assuming time-invariant baseline conditions, $\lambda_{0}(t)$ can be chosen as a constant, which gives the partial likelihood function as
\begin{equation}\label{eqn: likehood}
\begin{split}
    &f(E;G_{t},\boldsymbol{\theta})= C(\boldsymbol{\theta}) \prod_{e\in E}\lambda_{i_{e}j_{e}k_{e}}(t_{e}|G_{t_{e-1}},\boldsymbol{\theta}) \times \\
    &\exp{\left(-\Delta t_{e}\sum_{k=1,\cdots,K}\sum_{(i,j)\in D}\lambda_{ijk}(t_{e}|G_{t_{e-1}},\boldsymbol{\theta})\right)},
    \end{split}
\end{equation}
where $\Delta t_{e} = t_{e} - t_{e-1}$ is the time elapsed between the current event $e$ and the previous event $e-1$, and $C(\boldsymbol{\theta})$ is a normalization factor that makes this partial likelihood function well-defined. 
This expression represents the likelihood of every event in the sequence, along with the conditional likelihood that no other events occurred in the period between observed events. 


\subsection{Sufﬁcient Statistics}

The greatest utility of REM lies in the wide range of sufﬁcient statistics that can be derived from event sequences. 
These measures are numerical representations of speciﬁc interaction patterns, similar to those encoded in exponential random graph models (ERGMs) or stochastic actor-oriented models \cite{Lusher2013ExponentialNetworks}.
That is, like how conceptual models are translated into statistical models, sufﬁcient statistics are the operationalizations of model parameters including both endogenous and exogenous factors.

The endogenous factors summarize how the accumulation and sequencing of past relational events can inﬂuence the likelihood of the next relational event.
For instance, \textit{inertia} reﬂects the degree to which group members’ past contacts tend to be their future contacts;
\textit{Reciprocity} examines the likelihood of group member B sending a message to A if B had recently received a message from A.

Exogenous factors refer to characteristics outside the relational event history, including individual attributes, relational attributes, as well as environmental contextual information beyond the scope of the interaction system.
For example, sometimes the likelihood of a group interaction occurring is due to some \textit{individual attributes} of either the sender or receiver.
An extroverted individual might send messages at a higher rate than somebody who is more introverted.
\textit{Relational attributes} refer to different kinds of ties individuals might have with one another like afﬁnity (e.g., friendship, trust), ﬂow (e.g., other forms of sending messages), representational (e.g., endorsements), and semantic (e.g., shared interpretations) ties \cite{Shumate2013ANetworks}.
\textit{Environmental information} may encompass more readily measurable entities like the nature of the task, restrictions on communication channels, or availability of resources.

\section{Problem Formulation}\label{sec:problem formulation}

\begin{figure}
    \centering
    \includegraphics[width=0.48\textwidth]{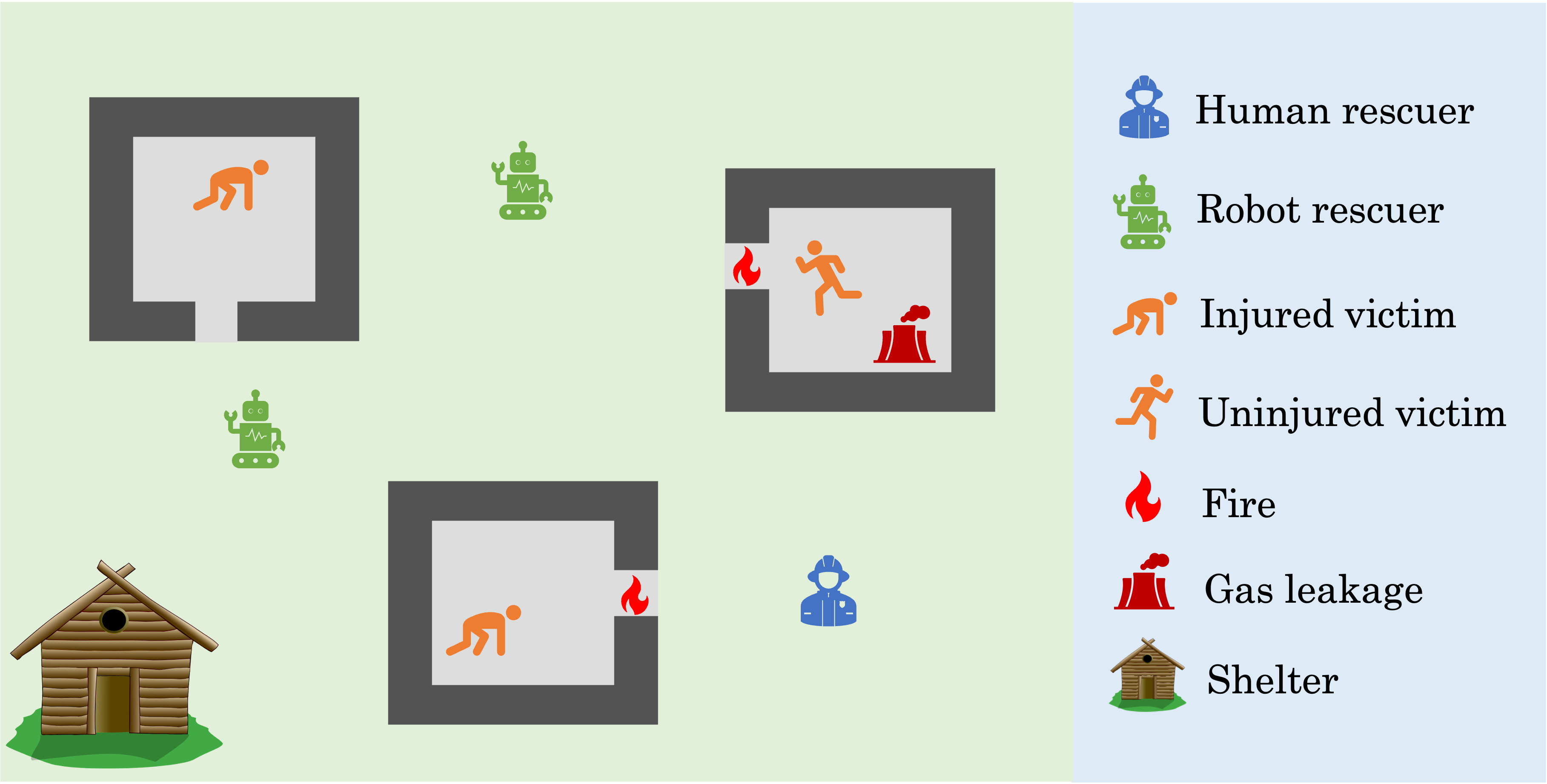}
    \caption{A human-robot collaborative SAR scenario where the human and multiple robots jointly perform the search and evacuation of victims from an incident area.}
    \label{fig:problem}
\end{figure}

We are inspired by a human-robot collaborative SAR scenario where the human and multiple robots jointly perform the search and evacuation of victims from an incident area \cite{Schoonderwoerd2022DesignTask}, as illustrated in Fig. \ref{fig:problem}.

\noindent \textbf{Environment:}
There are several buildings that have been hit by an earthquake, and whose entrances may be blocked by fire. 
Each building may contain one or more victims, who may be injured or uninjured.
One of the buildings is undergoing gas leakage inside, and if the gas density achieves a certain level, it will lead to an explosion.

\noindent \textbf{Task:}
The team needs to localize all victims in the buildings and assess their conditions. 
The injured victims should be treated and escorted to the shelter.

\noindent \textbf{Sub-task sequence dependencies:}
The victims blocked by fire cannot be accessed until the fire blocking the entrance is put down.
The injured victims must be treated before being transmitted to the shelter.

\noindent \textbf{Team member capabilities:}
The capabilities of the robots and humans are complementary.
Only robots can sense and estimate the location and density of the gas leakage.
Furthermore, they can put down fires and carry injured victims, but humans cannot.
In contrast, humans can treat injured victims, which cannot be done by robots.
However, both humans and robots can search and assess the victims, as well as shut down the gas leakage.
In this way, the task can only be achieved by the collaboration of different types of team members.

\noindent \textbf{Human-robot communications:}
Humans and robots can communicate through commands using template sentences.
The human has the option to either acquire the robots' status or give instructions to the robots.

The human can inquire about the status of robots by sending a command to a certain robot asking ``What are you doing?''
As a response, the inquired robot will provide the current sub-task that it is performing, for example, ``I'm going to Building X to put down the fire,'' or ``I'm escorting the injured victim to the shelter.''

The human can also instruct a certain robot by telling that robot to ``Go to Building X.''
If the instruction from the human conflicts with the robot's current sub-task, it can either obey the human's instruction regardless of its own mission by responding with ``Sure, I am going there.''; or stick to its current mission and convey it to the human ``Sorry, but I have to prioritize going to Building X.''

\noindent \textbf{Problem Statement:}
Based on the toy example scenario described above, the objective of this paper is to develop a trust-preserved shared autonomy framework that can: 1) dynamically infer the human's trust in the robots based on merely time-stamped commands communicated between them; and 2) adaptively coordinate the robot's autonomy level with the human's trust level, which facilitates both the completion of the task and the acceptability of the human.

\section{Dynamic Trust Inference}\label{sec:BREM}

\subsection{Quantification of Trust}

Following \cite{Mayer1995AnTrust}, we deﬁne trust as ``the willingness of a party to be vulnerable to the actions of another party based on the expectation that the other will perform a particular action important to the trustor, irrespective of the ability to monitor or control that other party''.
It is an essential component that is in general included in the time-varying attributes $G_{t}$. 
Since we are paying special attention to the trust between team members, we assume that the other attributes are known and static. 
For simplicity, in the rest of this paper, we denote $G_t$ to represent the unknown and dynamic trust that human places on the robots.

\subsection{Sufﬁcient Statistics}

\begin{figure}
    \centering
    \includegraphics[width=0.48\textwidth]{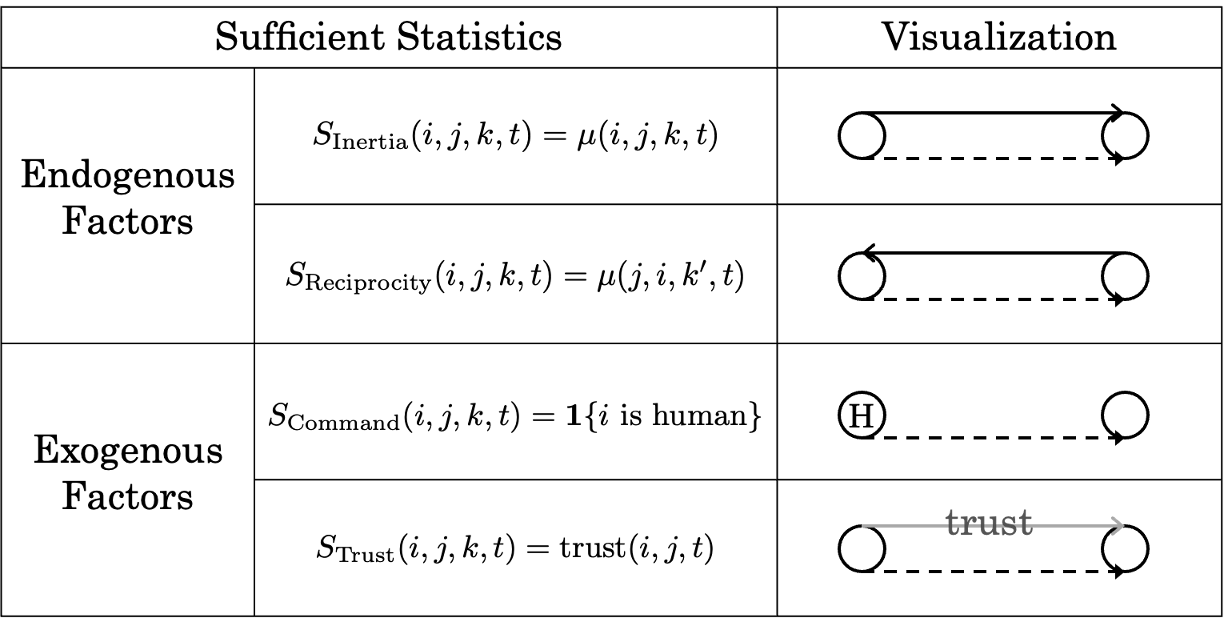}
    \caption{Examples of sufﬁcient statistics for the collaborative SAR scenario. The solid arrows represent past historical events, and the dashed arrows represent the future events to be predicted. $\mu(i,j,k,t)$ measures the frequency of event $(i,j,k)$ in the past event history before $t$.}
    \label{fig:SS}
\end{figure}

In the context of collaborative SAR scenario investigated in this paper, we utilize the sufﬁcient statistics described in Fig. \ref{fig:SS} to model the group communications. 
For endogenous factors, we include \textit{inertia} and \textit{reciprocity}.
For exogenous factors, we include the human's individual attribute of \textit{command} and the relational attribute of \textit{trust} between the human and the robots.
Those sufﬁcient statistics are by no means exhaustive, and various other structures may be encoded to capture speciﬁc behavioral patterns if needed.

\subsection{Bayesian Relational Event Modelling}\label{subsec:Bayesian REM}

\begin{figure}
    \centering
    \includegraphics[width=0.48\textwidth]{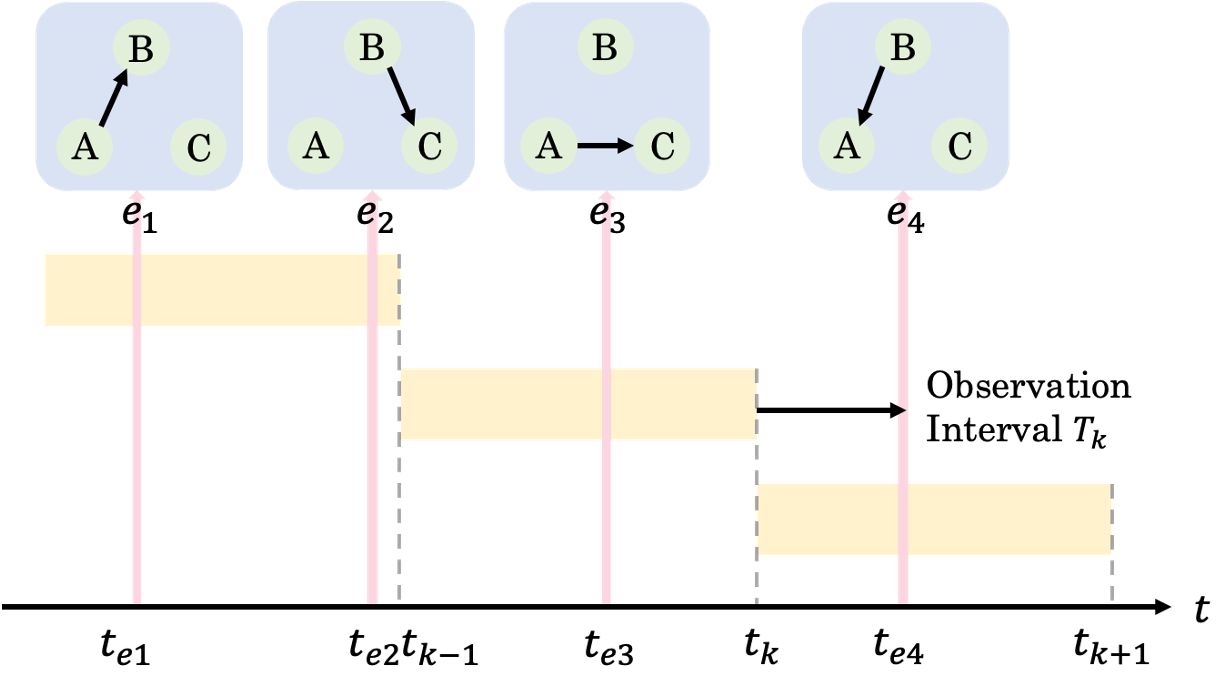}
    \caption{Bayesian REM framework. A moving window technique and sequential Bayesian update rule are utilized to update the estimation of team attributes online based on newly observed relational events. }
    \label{fig:BREM}
\end{figure}

To dynamically infer the evolution of human trust, we propose to advance REM into an adaptive framework taking advantage of Bayesian learning.
This framework is composed of two stages: 1) offline learning for team properties grounding, and 2) online learning for dynamic trust inference.

\noindent (1) \textbf{Team properties grounding.}
The sign and magnitude of each element in the coefficient parameter $\boldsymbol{\theta}$ determines how influential a particular network effect is regarding the generation of relational events across team members.
Therefore, the first stage aims to estimate the coefficient parameter $\boldsymbol{\theta}$ to ground the specific properties of the team to be investigated.
In this stage, a period of interaction history $E_{0}$ within the team is collected, as well as the initial trust information $G_{0}$ that can be obtained by a post-process survey.
Then the coefficient estimates for relational event models are derived from the solution of the maximum likelihood problem:
\begin{equation*}
    \boldsymbol{\theta}^{\ast} := \arg \max_{\boldsymbol{\theta}} \log \left( f(E_{0}; G_{0}, \boldsymbol{\theta}) \right).
\end{equation*}
Common optimization techniques such as Newton-Raphson are well-suited to solving this maximization problem. 

\noindent (2) \textbf{Dynamic trust inference.}
Based on the grounded properties $\boldsymbol{\theta}^{\ast}$ for a specific team, the next stage is to deploy a receding horizon approach to dynamically infer the time-varying trust that humans place on the robots, as illustrated in Fig. \ref{fig:BREM}. 
We assume that within each observation interval $T_{k}$  between two discrete observation time instant $t_{k-1}$ and $t_{k}$, the trust $G_{t}$ remains constant.
And the observed relational events history within the observation interval $T_{k}$ is recorded as $E_{k}$.
Define $\pi_{k}(G_{t})=P(G_{t}|E_{1:k}, \boldsymbol{\theta}^{\ast})$ as the posterior distribution of $G_{t}$ after $k$-th observation intervals. 
According to the \textit{Bayes' Theorem},
\begin{equation*}
\begin{split}
    \pi_{k}(G_{t})&=P(G_{t}|E_{1:k}, \boldsymbol{\theta}^{\ast})\\
    &=\frac{P(E_{k}|G_{t}, E_{1:k-1}, \boldsymbol{\theta}^{\ast})P(G_{t}|E_{1:k-1}, \boldsymbol{\theta}^{\ast})}{\sum_{G_{t}}P(E_{k}|G_{t}, E_{1:k-1}, \boldsymbol{\theta}^{\ast})P(G_{t}|E_{1:k-1}, \boldsymbol{\theta}^{\ast})} \\
    &=\frac{P(E_{k}|G_{t}, \boldsymbol{\theta}^{\ast})\pi_{k-1}(G_{t})}{\sum_{G_{t}}P(E_{k}|G_{t}, \boldsymbol{\theta}^{\ast})\pi_{k-1}(G_{t})} \\
    &=\frac{f(E_{k};G_{t}, \boldsymbol{\theta}^{\ast})\pi_{k-1}(G_{t})}{\sum_{G_{t}}f(E_{k};G_{t}, \boldsymbol{\theta}^{\ast})\pi_{k-1}(G_{t})},
\end{split}
\end{equation*}
where the likelihood function $f(E_{k};G_{t}, \boldsymbol{\theta}^{\ast})$ can be calculated according to \eqref{eqn: likehood}, and the prior distribution $\pi_{0}(G_t)$ offers the opportunity to take the initial trust information $G_{0}$ collected in the previous stage into consideration.

Superior to some point estimation approaches, e.g., maximum likelihood estimation (MLE), the utilization of Bayesian learning enables us to maintain the complete statistic description of trust through a probability distribution $\pi(G_t)$, which offers greater flexibility in quantifying trust level tailored to specific scenarios.
For example, the trust level $L_{\text{trust}}$ in general can be determined by taking expectation of the estimated trust distribution: $L_{\text{trust}} = \mathbb{E}_{\pi(G_t)}G_t$.
However, in some safety-critical scenarios, we may prefer to take a risk-averse perspective with certain risk measures: $L_{\text{trust}} = \mathcal{R}_{\pi(G_t)}G_t$, such as mean-variance, Value at Risk (VaR), and Conditional Value at Risk (CVaR) (cf. \cite{Wu2018AAsymptotics}).

\section{Trust-Preserved Shared Autonomy}\label{sec:trust-preserved SA}

\subsection{Quantification of Autonomy}

Level of robot autonomy, ranging from teleoperation to fully autonomous systems, defines the extent to which a system can carry out its own processes and operations without external control.
There have been different criteria for assessing the autonomy level, for example, the allocation between functions, or the amount of time that a person can neglect the robot \cite{Beer2014TowardInteraction}.
In the context of this paper, we quantify the autonomy level $L_{\text{autonomy}}$ of the robot as $L_{\text{autonomy}} = 1 - \alpha$, where $\alpha$ is the probability that the robot follows the human's instruction when contradicting its own reasoning.

\subsection{Coordination between Trust and Autonomy}

\begin{figure}
    \centering
    \includegraphics[width=0.48\textwidth]{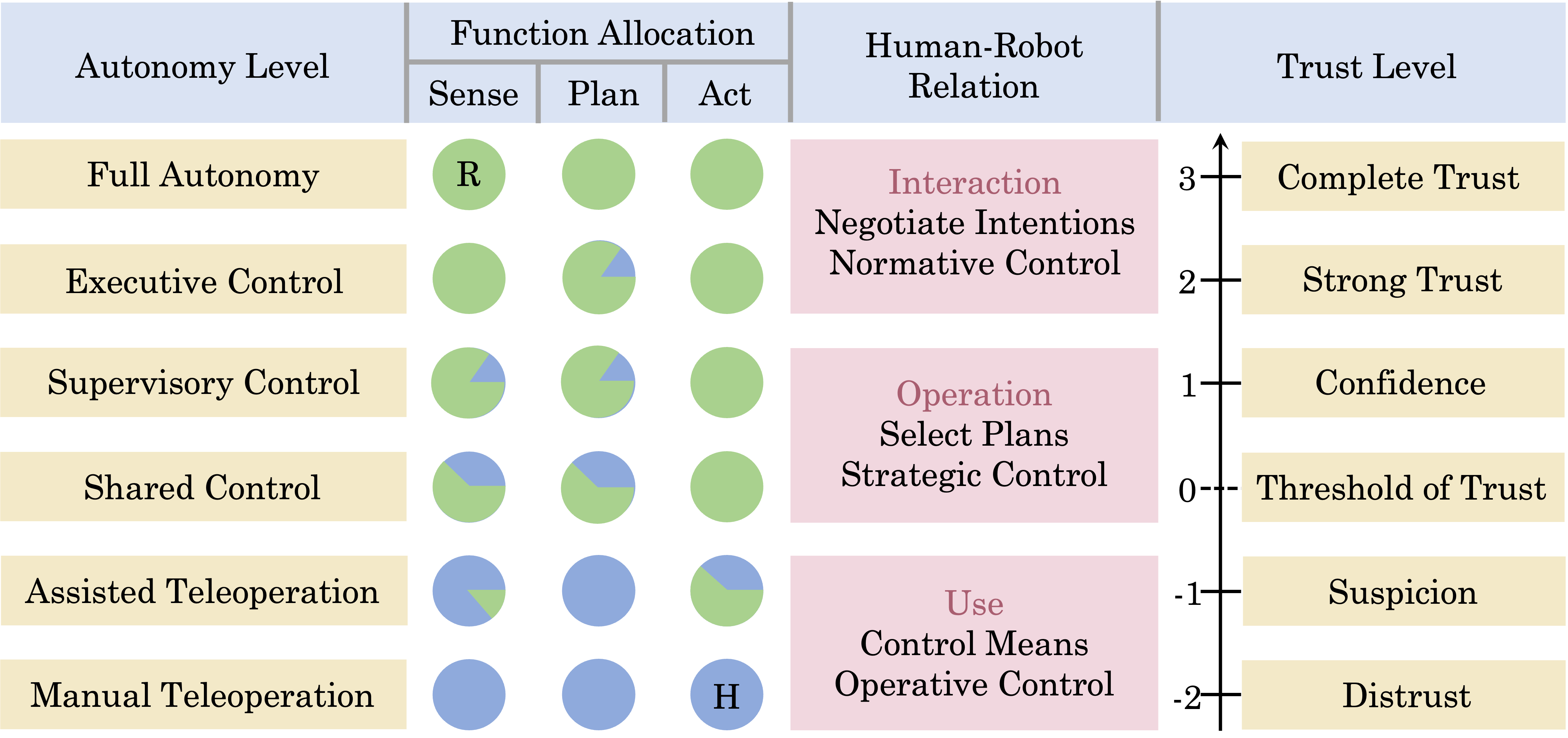}
    \caption{The coordination between the robot's autonomy level and the human's trust level.}
    \label{fig:autonomy_trust}
\end{figure}

Based on the inferred dynamics of trust level, the immediate question that arises is: How to effectively adjust the robot autonomy level according to the evolving human trust level?
In general, the more trust humans place on robots, the higher level of autonomy they tend to accept.
Therefore, an intuitive adaptive strategy is to passively choose the autonomy level that best matches the estimated trust level: $L_{\text{autonomy}} \propto L_{\text{trust}}$, as illustrated in Fig. \ref{fig:autonomy_trust}.
However, this reactive adjustment may impede the effectiveness and efficiency of the human-robot team and fail to take full advantage of the robot's capabilities.
For example, when humans maintain a lower level of trust in the robots, disuse or micromanagement may happen, which may impede the completion of task and increase the workload for human operators.

\subsection{Trust-preserved Shared Autonomy}

\begin{figure}
    \centering
    \includegraphics[width=0.48\textwidth]{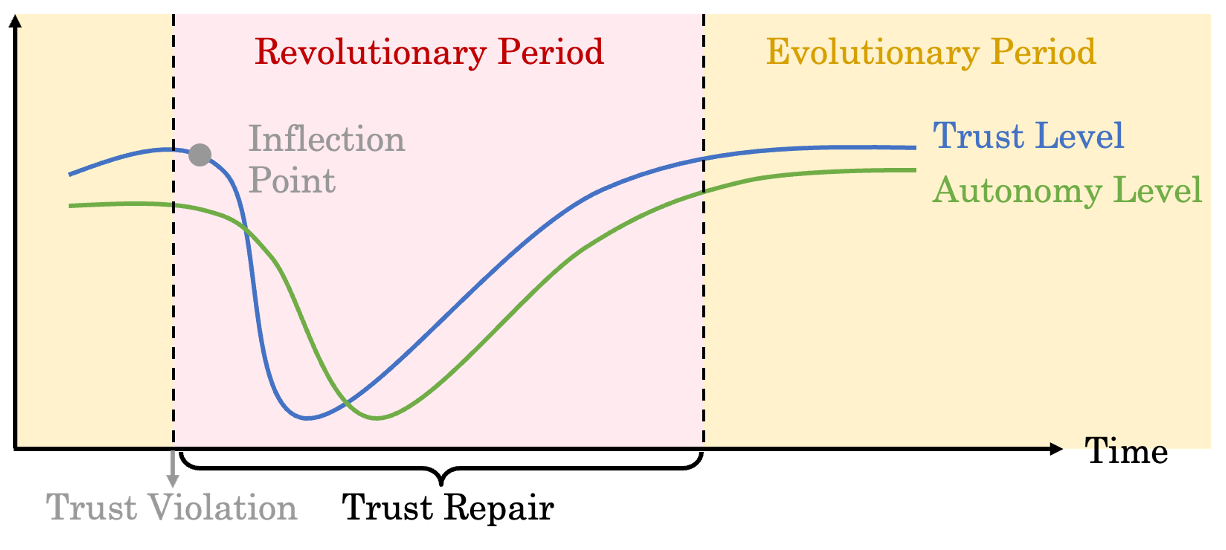}
    \caption{The punctuated equilibrium model of team dynamics, where the inflection point indicates a trust violation and the coming of revolutionary periods, during which some trust repair strategies should be deployed to re-establish the trust between humans and robots. }
    \label{fig:PEM}
\end{figure}

In real-world human-robot teaming scenarios, the collaboration between humans and robots is a long-term dynamic process, where all team members continuously engage in the process of establishing and calibrating trust among each other.
Therefore, robots are expected to deploy a trust-preserve strategy to actively establish, maintain, and repair human trust, rather than merely passively adapting to it.

On that account, we take a longitudinal perspective of human-robot teaming and analyze the temporal dynamics of trust formation and maintenance. 
Punctuated Equilibrium Model (PEM) \cite{Sabherwal2001TheModel}, which describes the pattern of evolution in evolutionary biology and paleontology, reveals that the dynamics of groups often involve long periods of relative stability (equilibrium) punctuated by relatively short bursts of rapid change (punctuation), as illustrated in Fig. \ref{fig:PEM}.
According to the paradigm of PEM, the team structure is considered stable and experiences limited change over the \textit{evolutionary periods}, representing equilibrium.
These periods are interrupted by brief but significant episodes of \textit{revolutionary periods}, where the team structure may disassemble, reconfigure, and experience qualitative, metamorphic change.
In the context of human-robot teaming, \textit{evolutionary periods} represent the periods when humans and robots have established and maintained a relatively stable trust and relationship.
Within those periods, robots can simply adjust the autonomy level to match the estimated trust level.
However, for \textit{revolutionary periods}, which often happen at the beginning of the teaming process or when trust violation occurs, additional trust repair approaches should be deployed to establish or re-establish the trust between team members.

\begin{table*}[t]
    \centering\normalsize
    \begin{tabular}{| m{4cm} || m{10cm}|}
        \hline
       Methods for trust repair  &  Description \\
       \hhline{|=#=|}
       Apology  & Admission of fault and statement of remorse \\
       \hline
       Denial & Recognition of error and assertion that they are not at fault \\
       \hline
       Control & Statement of control/awareness of situation \\
       \hline
       Convey uncertainty directly & Conveying uncertainty directly by providing machine conﬁdence indicators \\
       \hline
       Show critical states & Enhance an operator’s mental state by showing only the critical states, situations that require immediate action, of an algorithm’s policy \\
       \hline
    \end{tabular}
    \caption{Typical trust repair strategies suitable for remote human-robot collaboration scenarios.}
    \label{tab:trust repair}
\end{table*}

The transition from an evolutionary period into a revolutionary period is characterized by the \textit{inflection point} of the team dynamics, which can be detected by inspecting the derivatives of the estimated trust level. 
Whenever a revolutionary period is detected, various trust repair approaches, including explaining the cause of a mistake, or making promises about future behavior, can be actively deployed.
Following \cite{DeVisser2020TowardsTeams}, we summarize several strategies suitable for human-robot collaboration SAR scenarios, as demonstrated in Table \ref{tab:trust repair}.
For instance, robots could explain their decisions to the human (e.g., ``My priority has been adjusted due to an observed dangerous condition in Building X.''), or seek confirmation from human whenever making a counter-command decision (e.g., ``Instruction received. However, I have the following additional environmental information that may be useful for you to refine the decisions. Please resend your decision.'').

\section{User Study}

To verify whether the proposed trust-preserved shared autonomy approach actually assists a human operator in executing tasks that require a high workload and fluent human-robot collaboration, a user study was conducted on a human-robot collaborative SAR scenario.

\subsection{Study Design}

\noindent \textbf{Participants:} 
We recruited 16 college student subjects (3 female, average ages 26.06$\pm$4.93 years) to participate in our study. 
All subjects were provided informed consent prior to the experiment.

\noindent \textbf{Tasks:} 
Each participant needs to play a simulated human-robot collaborative SAR game, as described in Section \ref{sec:problem formulation}.

\begin{figure}
    \centering
    \includegraphics[width=0.5\textwidth]{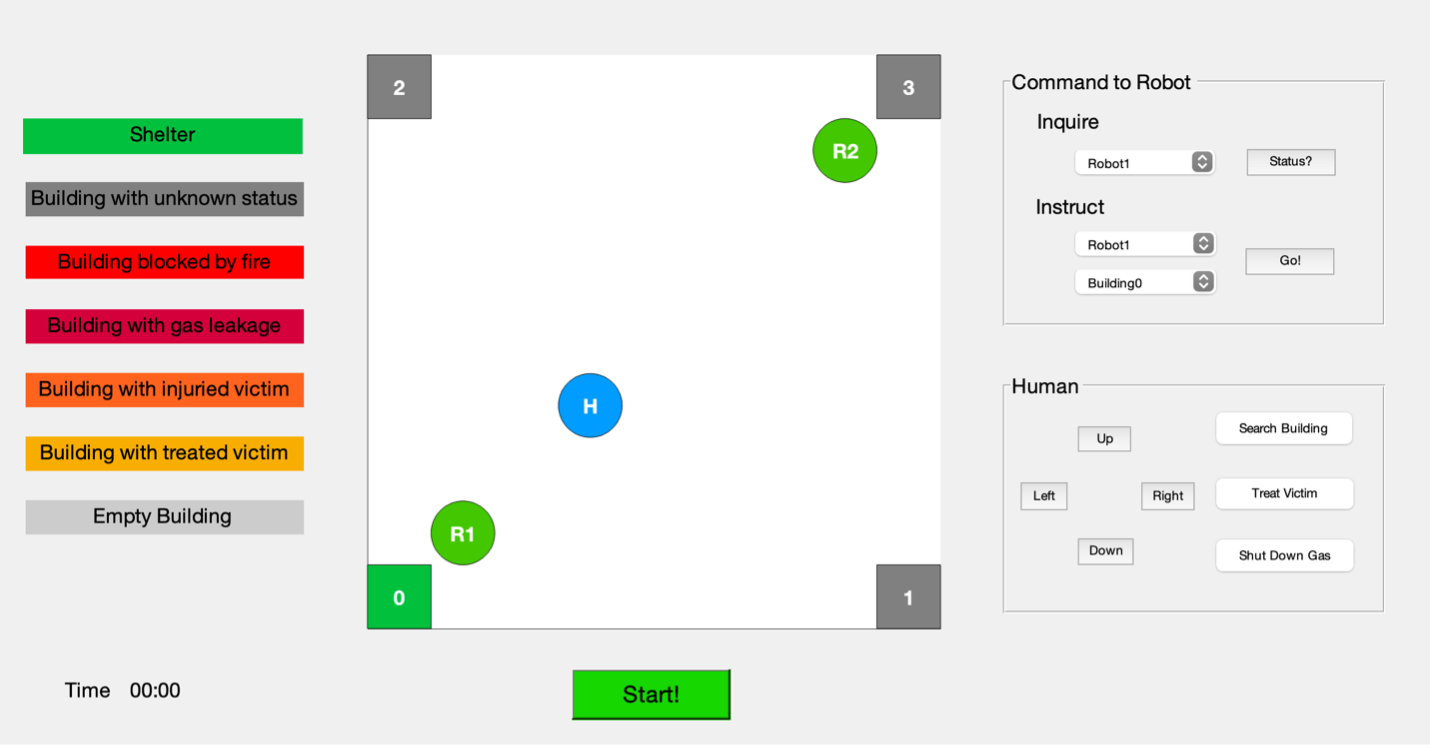}
    \caption{The simulation interface used by the user study to perform the SAR task. The center figure displays the workspace of SAR. The icons on the left of the workspace demonstrate the colors that are used to represent different statuses of those buildings. The right interface part displays the actions that the user can conduct. }
    \label{fig:interface}
\end{figure}

\noindent \textbf{Experimental Setup:} 
A simulation interface to perform the SAR task was developed using the Graphical User Interfaces (GUIs) tool in Matlab, as illustrated in Fig. \ref{fig:interface}.
The center figure displays the workspace of SAR, where the buildings are represented with squares. 
We deploy a team that contains one human and two robots, whose locations are displayed by the blue circle ``H'' and green circles ``R1'' and ``R2'', respectively.
The icons on the left of the workspace demonstrate the colors used to represent the different statuses of those buildings. 
The right interface part displays the actions that the user can conduct. 
In the ``Command to Robot'' panel, users can inquire about about each robot’s current targeted building by choosing the corresponding robot in the pop-menu and hitting the “Status?” button.
Similarly, users can also instruct a certain robot to go to a certain building by changing the pop-menus and hitting the “Go!” button.
The ``Human'' panel is used by the users to control the human's movement, and conduct different sub-tasks.

\noindent \textbf{Procedure:} 
Each participant received an instruction document that contained an explanation of the scenario and tasks, and an instruction on how to use the interface.

The participant first played the game in the practice phase to get familiar with the task and interface. 
In this phase, the robots were assigned two fixed autonomy levels, one with a higher autonomy level ($L_{\text{autonomy}}=0.9$) and one with a lower autonomy level ($L_{\text{autonomy}} = 0.1$).
To prevent the participants from learning how to go about the entire SAR operation, the scenarios (which building was burning, which one had a gas leak, where they were located, which one had people) were randomized for each trial.
Participants could try as many trials as they wish at this stage, after which they were asked to report their 11-point Likert scale on their trust in each robot (in what degree they tend to accept the robot's decisions).
The interaction and trust data collected at this stage was utilized to perform the offline learning for team properties grounding, which captured the individual differences of the participants (e.g. if a higher trust predicts the probability of sending fewer inquiring messages).

In the next stage, a within-subject design was applied, namely, each participant completed the tasks with two conditions (\textit{Baseline SA} and \textit{Trust-preserved SA}) for 6 trials, with the order of conditions counterbalanced across the participants. 
After completing the experimental tasks, participants were administered a brief questionnaire to assess their user experiences.

\noindent \textbf{Variables:} 
For comparison, we tested two conditions with different settings of robot autonomy.
In the baseline condition (\textit{Baseline SA}), each robot follows a fixed autonomy level according to the reported trust at the first stage.
On the contrary, the robots in another condition (\textit{Trust-preserved SA}) are equipped with our proposed strategy, which are capable of dynamically inferring human trust and seamlessly coordinating their autonomy level during the operation.

\subsection{Measures}

\noindent \textbf{Objective Measures:} 
Defining a successful task execution as completing the task without a gas explosion, we recorded each participant's \textit{success rate} over multiple trials with both conditions.
For each successful task execution, performance measures were obtained by automatically logging data during task execution. 
The following measures were collected: \textit{task duration} (shorter completion time means better performance);
and the \textit{number of commands} sent by the human agent (fewer commands means better performance).

\noindent \textbf{Subjective Measures:} 
We administered an 11-point Likert scale (from ``0-Not at all agree'' to ``10-Totally agree'') survey for each condition, which is a tailored version of \cite{Hoffman2019EvaluatingCollaboration} within the SAR scenario.
The test distinguishes five dimensions of collaboration fluency: How \textit{easy} it was to complete the tasks; How \textit{helpful} the robots were to complete the tasks; How much \textit{trust} had been put into the robots about their situation awareness; In what degree the robots acted like a \textit{member} of the team; In what degree the robots' responses helped \textit{improve} the trust.
We also asked each participant's \textit{preference} on which condition they prefer to work with in the future.

\subsection{Results}

\begin{figure}
    \centering
    \includegraphics[width=0.5\textwidth]{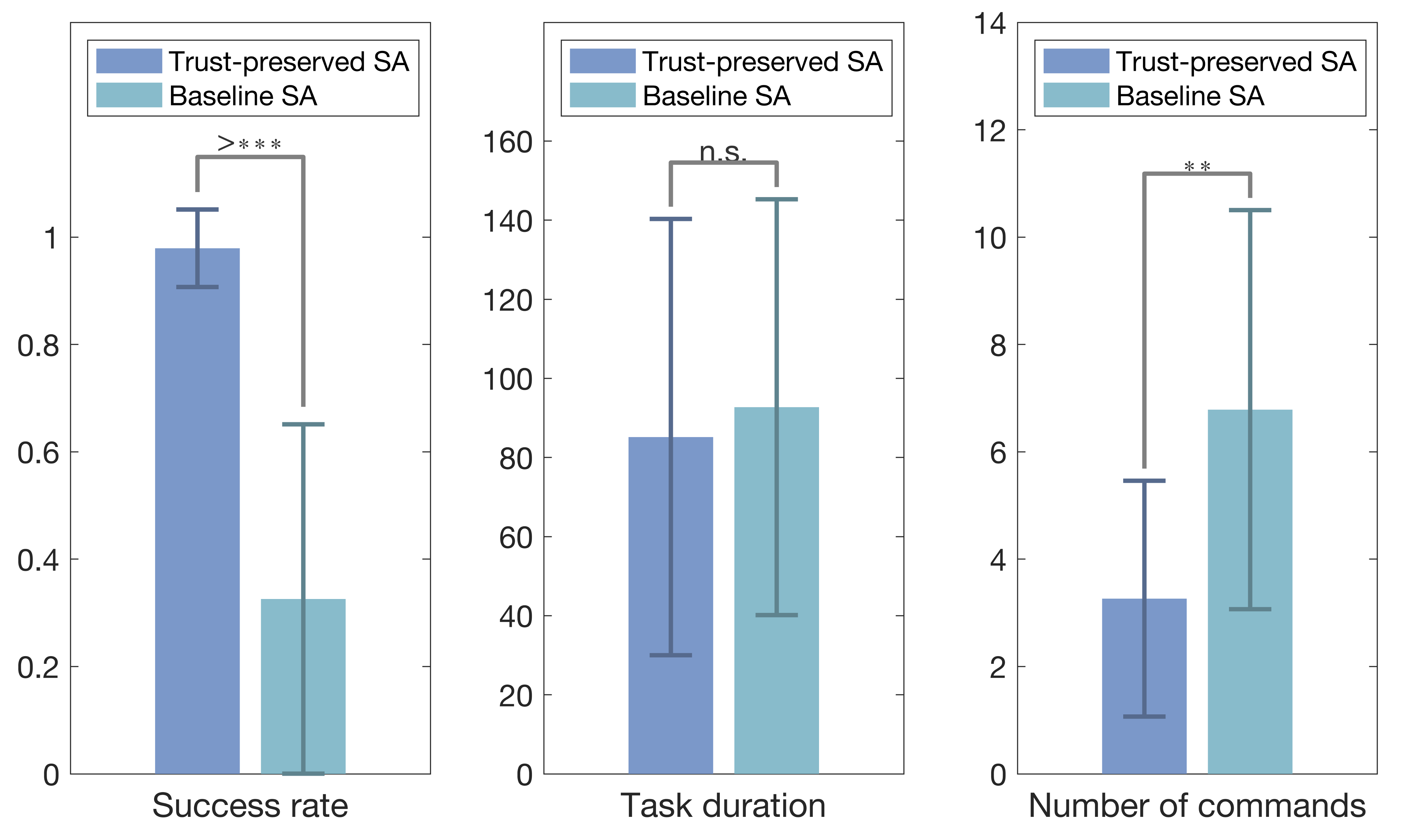}
    \caption{Results of objective measures.}
    \label{fig:res_obj}
\end{figure}

\begin{figure}
    \centering
    \includegraphics[width=0.5\textwidth]{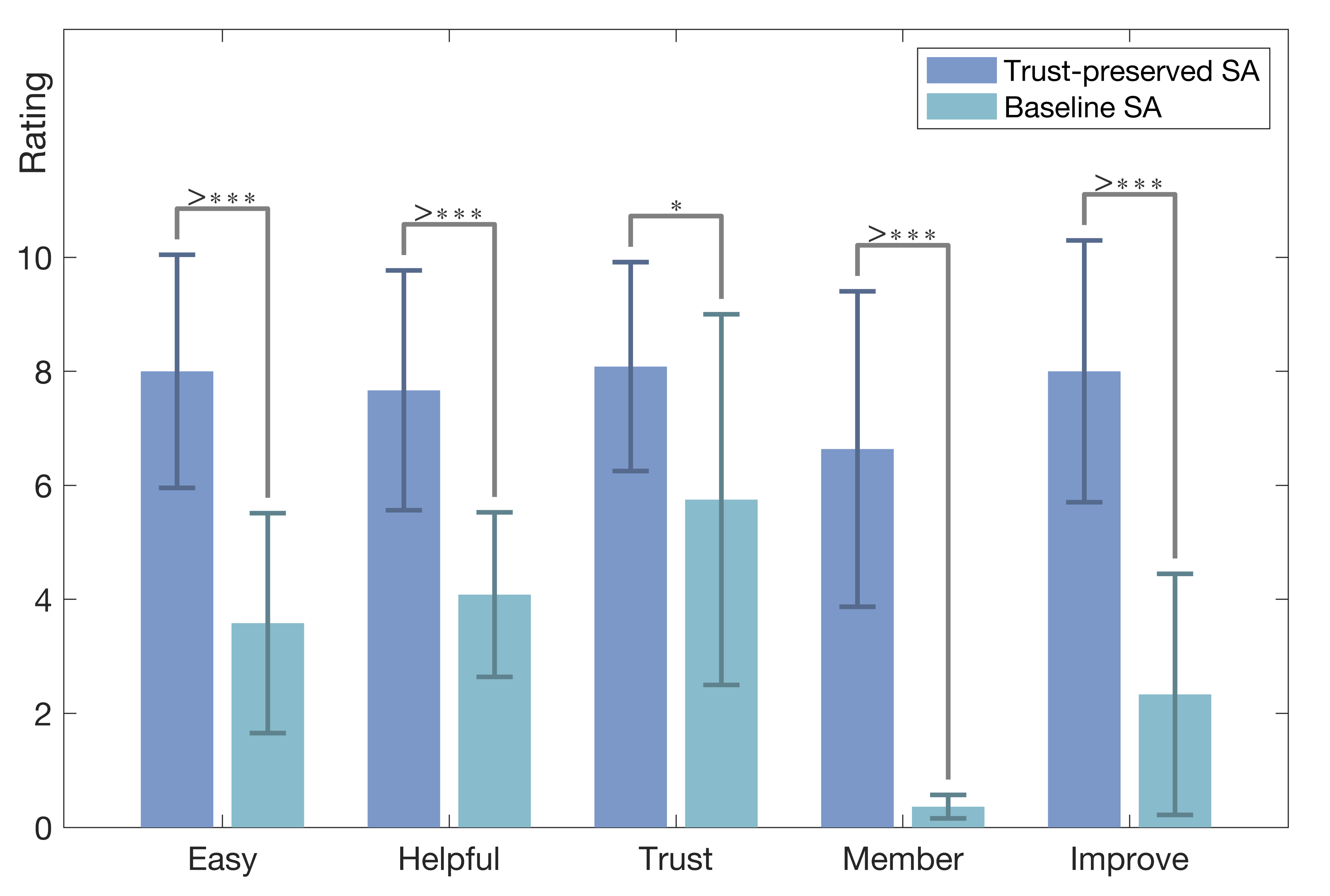}
    \caption{Results of subjective measures. }
    \label{fig:res_sub}
\end{figure}

The means, standard deviations, and t-test results (where n.s., $\ast$, $\ast\ast$, $\ast\ast\ast$, $>\ast\ast\ast$ denote $p> 0.05, \le 0.05, \le 0.01, \le 0.001, \le 0.0001$, respectively) of objective and subjective measures are displayed in Fig. \ref{fig:res_obj} and Fig. \ref{fig:res_sub}. 

\noindent \textbf{Objective Results:} 
As illustrated in Fig. \ref{fig:res_obj}, the proposed \textit{Trust-preserved SA} approach significantly increased the \textit{success rate} compared to the \textit{Baseline SA} approach, $t(30)=6.41, p<.00001$, achieving 98\% average success rate. The effect size was medium, with a Cohen’s d of 0.7.
In addition, it can also reduce the workload of humans by effectively decreasing the \textit{number of commands} sent from the human teammate, $t(30)=-2.62, p=.009$. The effect size was very small, with a Cohen’s d of 0.05.

\noindent \textbf{Subjective Results:}
From Fig. \ref{fig:res_sub}, it is obvious to interpret significant improvements in the users' experience over all five dimensions of collaboration fluency.
Furthermore, when asked to provide their preferences of conditions to work with in the future, all 16 participants chose our proposed \textit{Trust-preserved SA} method.

\subsection{Discussion}

Those results were consistent with our hypothesis that, the proposed trust-preserved shared autonomy strategy can improve the performance of human-robot teaming in terms of both objective measures about task completion, and subjective measures about the human acceptability.

During this study, we observed that humans tended to place low levels of initial trust in robots at the beginning of the task. 
And the reason that our proposed strategy out-performed the baseline may due to its ability to detect human trust and actively deploy trust repair approaches when necessary.
This enabled robots to continuously operate with a high level of autonomy to better facilitate the overall team performance.
The reason that there is no significant difference in the \textit{task duration} between the two conditions may be because only the completion time of successful task execution was recorded. 
Due to the existence of time limit about gas explosion, task duration larger than certain threshold was recognized as failure and will not be recorded.

Although primitive, the proposed approach and designed user study within the SAR scenario provide a valuable first step to develop trust inference and trust-preserved shared autonomy for remote human-robot teaming. 
After the user study, we asked the participants to provide their suggestions to enhance the scenario setup. 
Although 14 of them were satisfied with the design, 2 of them suggested to provide an option of human manual override or disuse, which could ensure human authority when needed.
Therefore, one interesting next step is to extend this basic scenario setup to more general and complex human-robot teaming scenarios, for example, considering the disuse of robots or dynamically changing priorities.

\subsection{Limitations}
The 16 participants in our user study were college students and had no prior experiences on real-world SAR. By contrast, experts with professional training on SAR may have biased expectations and assessments of the robot performance. 
In addition, the real SAR scenario is far more complicated than the one emulated here, which may lead to different results.

\section{Conclusion and Future Work}

In this paper, a trust-preserved shared autonomy strategy was proposed, which allows robots to seamlessly adjust their autonomy level, striving to optimize team performance and enhance their acceptance among human collaborators.
Providing an innovative approach to model the dynamics of human trust based on Bayesian REM, this paper opens a new perspective of trust modeling in remote human-robot interaction scenarios.
Future work plans to validate the proposed method using experiments performed with real robots and realistic interfaces.

\bibliographystyle{unsrt}
\bibliography{references}

\end{document}